\documentclass{article}

\usepackage{microtype}
\usepackage{graphicx}
\usepackage{subcaption}
\usepackage{booktabs} % for professional tables
\usepackage[dvipsnames]{xcolor}      % 用于 ForestGreen 和 magenta 颜色
\usepackage{pifont}                 % 用于提供 \checkmark 和 \xmark 的基础符号 (可选，见下方说明)
\usepackage{amssymb}                % 常用符号包
\usepackage{makecell}               % 用于表格单元格内换行 (\makecell)
\usepackage{makecell}               % 提供 \Xhline 这种加粗横线
\newcommand{\xmark}{\ding{55}}
\usepackage{colortbl}
\usepackage{multirow} 
\usepackage{hyperref}

\usepackage[preprint]{icml2026}

\usepackage{amsmath}
\usepackage{amssymb}
\usepackage{mathtools}
\usepackage{amsthm}
\usepackage[capitalize,noabbrev]{cleveref}

\theoremstyle{plain}

\theoremstyle{definition}

\theoremstyle{remark}

\usepackage[textsize=tiny]{todonotes}

\icmltitlerunning{CrystaL: Spontaneous Emergence of Visual Latents in MLLMs}

\begin{document}

\twocolumn[
  \icmltitle{CrystaL: Spontaneous Emergence of Visual Latents in MLLMs}



  \icmlsetsymbol{equal}{*}
  \icmlsetsymbol{corresp}{$\dagger$}
  
  \begin{icmlauthorlist}
    \icmlauthor{Yang Zhang}{nankai,equal}
    \icmlauthor{Danyang Li}{nankai,equal}
    \icmlauthor{Yuxuan Li}{nankai,corresp}
    \icmlauthor{Xin Zhang}{nankai}
    \icmlauthor{Tianyu Xie}{xiada}
    \icmlauthor{Mingming Cheng}{nankai}
    \icmlauthor{Xiang Li}{futian,nankai,corresp}
  \end{icmlauthorlist}

  \icmlaffiliation{nankai}{VCIP, School of Computer Science, Nankai University}
  \icmlaffiliation{futian}{Nankai International Advanced Research Institute (Shenzhen Futian)}
  \icmlaffiliation{xiada}{Xiamen University}

  \icmlcorrespondingauthor{Yuxuan Li}{yuxuan.li.17@ucl.ac.uk}
  \icmlcorrespondingauthor{Xiang Li}{xiang.li.implus@nankai.edu.cn}

  \icmlkeywords{Machine Learning, ICML}

  \vskip 0.3in
]



\printAffiliationsAndNotice{\icmlEqualContribution}  % no special notice (required even if empty)

\begin{abstract}

Multimodal Large Language Models (MLLMs) have achieved remarkable performance by integrating powerful language backbones with large-scale visual encoders. Among these, latent Chain-of-Thought (CoT) methods enable implicit reasoning in continuous hidden states, facilitating seamless vision–language integration and faster inference. However, existing heuristically predefined supervision signals in latent CoT provide limited guidance for preserving critical visual information in intermediate latent states. To address this limitation, we propose CrystaL (Crystallized Latent Reasoning), a single-stage framework with two paths to process intact and corrupted images, respectively. By aligning the attention patterns and prediction distributions across the two paths, CrystaL crystallizes latent representations into task-relevant visual semantics, without relying on auxiliary annotations or external modules. Extensive experiments on perception-intensive benchmarks 
% (CVBench, HRBench, VStarBench) 
demonstrate that CrystaL consistently outperforms state-of-the-art baselines, achieving  gains in fine-grained visual understanding while maintaining robust reasoning capabilities. Codes are available \href{https://github.com/yangzhangok/crystal}{here}.

\end{abstract}

\section{Introduction}

\begin{figure*}[h]
    \centering
    \includegraphics[width=1\linewidth]{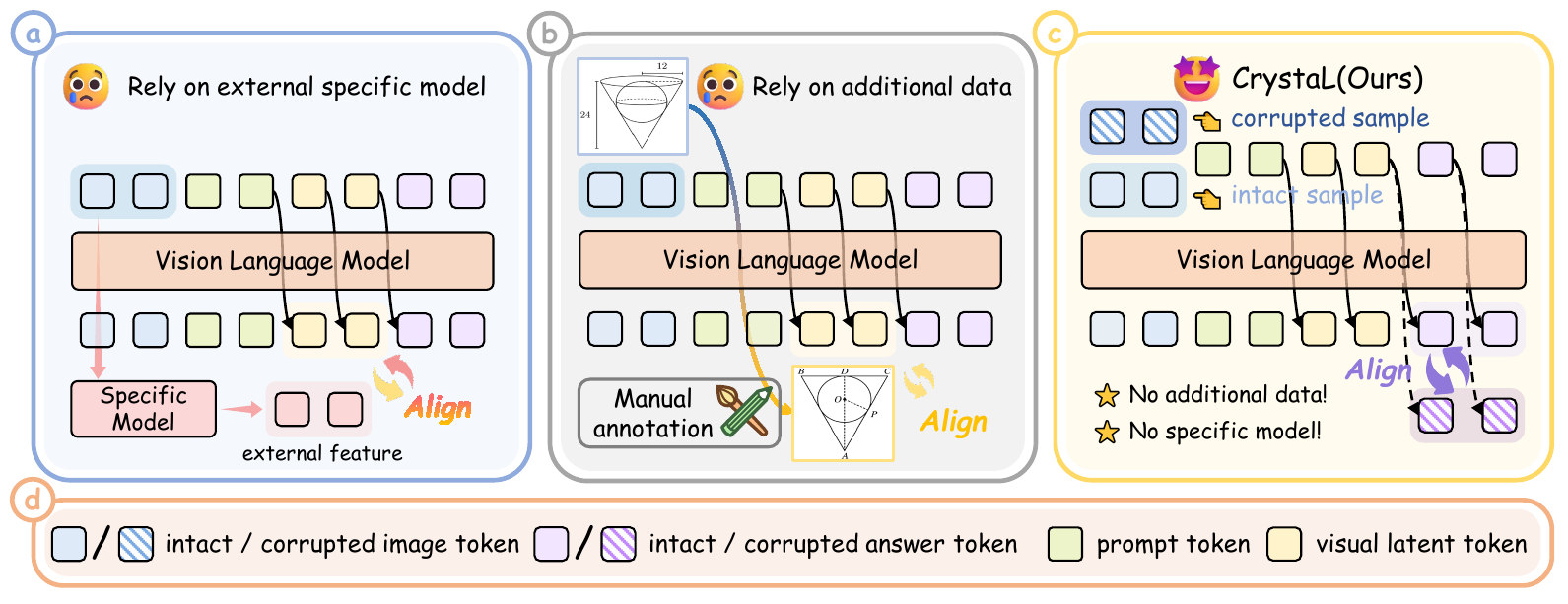}
    \caption{\textbf{Previous Paradigms vs. Our Paradigm (CrystaL).} (a) describes the paradigm of supervising visual latent tokens via predefined features from specific models, such as SAM~\cite{kirillov2023segment}, DINO~\cite{caron2021emerging}. (b) denotes the process of modifying original image to guide the reasoning steps. (c) While other methods use auxiliary modules or data to train the visual latent token, CrystaL can do it by self-supervising in a single-stage.}
    \vspace{-12pt}
    \label{fig:contrast}
\end{figure*}

The rapid evolution of Multimodal Large Language Models (MLLMs)~\cite{liu2023visual} has significantly bolstered general-purpose vision-language understanding through the integration of large-scale visual encoders and potent language backbones. By scaling model capacity and diversifying multimodal training corpora, recent systems~\cite{chen2024internvl,bai2025qwen2,wang2024qwen2,yang2025qwen3,zhu2025internvl3} have achieved remarkable performance across a broad spectrum of tasks~\cite{zhang2025unichange,li2026wowseg,zhao2025NAIPv1,zhao2025naipv2}. To further cultivate complex reasoning, many MLLMs incorporate explicit reasoning mechanisms inspired by Chain-of-Thought (CoT)~\cite{zhang2023multimodal,shao2024visual}, which partition intricate queries into intermediate inferential steps.

% Within this context, three predominant paradigms for multimodal CoT have emerged. Textual CoT ~\cite{xu2025llava} converts visual content into discrete linguistic rationales; while highly interpretable, it often suffers from substantial information loss during the quantization from continuous visual signals to discrete tokens. Tool-augmented CoT ~\cite{zheng2025deepeyes} invokes specialized external modules via dedicated tokens to generate structured traces, improving precision at the cost of significant pipeline dependencies. Conversely, Latent CoT approaches ~\cite{chen2025mint,gao2025interleaved,yang2025machine} perform reasoning implicitly via continuous hidden states, thereby bypassing the constraints of symbolic representations. These methods not only accelerate inference compared to textual and tool-based counterparts but also facilitate a more seamless integration of vision and language. Consequently, this paradigm has catalyzed promising advancements ~\cite{su2025patch,zhang2025latent} in interleaved text-image reasoning.

There are three main paradigms that exist for multimodal CoT, distinguished by their reasoning space. Textual CoT~\cite{xu2025llava} converts visual inputs into discrete natural language rationales, while tool-augmented CoT~\cite{zheng2025deepeyes} extends this space with dedicated tokens to invoke external modules. In contrast, latent CoT~\cite{chen2025mint, gao2025interleaved, yang2025machine} performs reasoning directly in continuous hidden states, preserving visual information, avoiding quantization loss, and enabling seamless integration of vision and language. This leads to faster inference and more coherent multimodal reasoning, driving recent advances in interleaved text–image tasks~\cite{su2025patch, zhang2025latent}.

% Despite their flexibility, these hidden states inherently lack the explicit guidance in textual rationales from the vanilla next token loss to ensure the preservation of critical visual evidence from the image, such that the high-dimensional latent space often fails to anchor itself to essential visual cues. Current research endeavors to bridge this gap through costly human-annotated trajectories~\cite{wang2025monet}, dense supervision derived from auxiliary vision models~\cite{qin2025chain}, or the integration of additional architectural modules~\cite{yu2025vismem}. They intricately design the objective loss. While contemporary efforts such as ~\cite{li2025latent_livr} recognize this gap, they primarily address it through heuristic architectural constraints—specifically by masking attention patterns to create visual bottlenecks. However, such structural interventions often necessitate a complex multi-stage training pipeline and may not fully guarantee the functional utility of latent tokens for the final generative task. 
Despite these advantages, early latent CoT methods face a limitation: existing supervision signals are often not fully aligned with the nature of latent reasoning. In contrast to textual CoT, where the next-token prediction objective explicitly constrains each intermediate reasoning step, latent CoT relies on supervision that is only indirectly related to the underlying reasoning process. Specifically, the standard next-token loss provides supervision only at the final output level, offering little guidance on how intermediate latent states should encode or preserve critical visual evidence.

To compensate for this mismatch, prior works introduce auxiliary supervision from external vision models such as CoVT~\cite{qin2025chain} or human-annotated images such as Monet~\cite{wang2025monet}, which is at the price of external modules. However, such signals are heuristic and task-dependent, as they enforce similarity to pre-defined visual features rather than directly supervising reasoning fidelity. Although recent work such as LIVR~\cite{li2025latent_livr} attempts to address this problem via enforcing visual bottlenecks through attention masking, such approaches require multi-stage training pipelines and still cannot guarantee that latent tokens are functionally aligned with the final generation objective. Consequently, such regularization fails to ensure that the latent space maintains semantically grounded visual information. This limitation indicates that the challenge lies not in the absence of supervision, but in the inadequacy of existing supervision paradigms to effectively guide latent reasoning.

In contrast, we propose \textbf{CrystaL} (Crystallized Latent Reasoning), a principled \textit{single-stage} framework designed to resolve the misalignment between existing supervision signals and latent reasoning, through a consistency-driven training paradigm. Unlike prior works that necessitate complex multi-stage pipelines, CrystaL employs a dual-path architecture comprising a \textit{intact path} and a \textit{corrupted path}. Specifically, the intact path undergoes standard multimodal training on intact images, while the corrupted path processes visually degraded counterparts. Visual latent representations are copied from the intact path to the corrupted one, enabling the model to recover the ground-truth answer despite the corrupted input. To ensure that these latents are functionally indispensable, we minimize the divergence between the paths' cross-path attention patterns and output distributions. By enforcing such cross-path consistency, we guide the latent states to ``crystallize'' into targeted visual semantics which compensate for the corrupted images.

By virtue of this consistency-driven objective, CrystaL encourages latent representations to anchor targeted visual information in a fully self-supervised and task-aligned manner, effectively eliminating the dependence on external priors or manual annotations. Empirical evaluations on perception-intensive benchmarks (including CVBench~\cite{tong2024cambrian}, HRBench~\cite{wang2025divide}, and VStarBench~\cite{wu2024v}) demonstrate that CrystaL yields substantial gains in fine-grained visual understanding while maintaining robust reasoning capabilities. Specifically, CrystaL achieves 76.6\% on 2D CVBench, 84.4\% on 3D CVBench, 73.4\% on 4K HRBench, and 71.1\% on 8K HRBench, surpassing CoVT and LIVR by substantial margins, while also yielding the highest average score (75.4\%) across all benchmarks (See Table~\ref{tab:main}). These results suggest that CrystaL provides a scalable and semantically grounded solution to the long-standing challenge of misalignment between supervision signals and visual latent reasoning. By ensuring that latent inferential steps remain coupled with visual evidence, CrystaL facilitates a more reliable multimodal intelligence.

% In summary, this work makes three main contributions:
% \begin{itemize}
%     \item We identify a critical bottleneck in visual latent reasoning: the absence of intrinsic learning signals that explicitly enforce visual grounding during the inferential process.
%     \item We propose CrystaL, a single-stage latent CoT framework that introduces a dual-path consistency objective to provide task-driven supervision for latent states without auxiliary labels.
%     \item Through extensive experiments, we demonstrate that CrystaL significantly enhances visual fidelity and reasoning performance, surpassing both direct fine-tuning and state-of-the-art baselines across multiple perception-heavy benchmarks.
% \end{itemize}

In summary, this work makes the following contributions:

1. We propose \textbf{CrystaL}, a principled single-stage latent CoT framework that guides latent representations to capture task-relevant visual information, enabling efficient and coherent multimodal reasoning.

2. We introduce a dual-path consistency objective that provides task-aligned self-supervision for latent representations, independent of auxiliary annotations or external modules.

3. Extensive experiments demonstrate that CrystaL substantially improves vision-dependent reasoning performance across multiple perception-intensive benchmarks, outperforming strong baselines while maintaining efficient training and inference.

\section{Related Work}

\subsection{Multimodal Chain-of-Thought Reasoning}
% 受纯语言模型中思维链方法成功经验的启发，近期研究致力于赋予MLLMs类似的多步推理能力。早期基于文本的CoT方法通过提示使模型在给出最终答案之前生成显式的推理过程~~\cite{zhang2023multimodal, amazon2024mcot, xu2025llava, apple2025cot}。尽管这些方法在科学问答和数学问题上表现有效~~\cite{lu2024mathvista}，但它们仍受限于离散语言符号在描述连续视觉细节时的表达能力不足。模型倾向于依赖自身生成的文本，而非原始视觉输入，从而导致信息丢失和幻觉现象~~\cite{liu2024hallucination, deepthinking2025}。
% 为解决这一问题，DeepEyes~~\cite{zheng2025deepeyes} 和 Visual Programming~~\cite{gupta2023visualprog} 等工具增强型框架将感知任务交由外部视觉模型处理。然而，这类流水线不可微分，且严重依赖现成工具的质量。近期的一些混合方法尝试通过将视觉特征与文本交错融合来实现推理的视觉锚定~~\cite{yang2025mirage}，但它们通常需要昂贵的人工标注推理轨迹或边界框监督。与这些方法不同，我们提出的 CrystaL 框架培养了一种内在的推理能力，该能力完全可微且自监督，无需依赖外部工具或密集标注。
Inspired by the success of CoT approaches in pure language models, recent research has focused on endowing MLLMs with similar multi-step reasoning capabilities. Early text-based CoT methods employed prompts to guide models in generating explicit reasoning steps before providing final answers~\cite{zhang2023multimodal, zhang2025improve, xu2025llava}. Whilst effective for scientific Q\&A and mathematical problems~\cite{lu2023mathvista}, these approaches remain constrained by the limited expressiveness of discrete linguistic symbols when describing continuous visual details. Models tend to rely on self-generated text rather than raw visual inputs, leading to information loss and hallucination phenomena~~\cite{wu2025grounded, dong2025mirage}.

To address this, tool-augmented frameworks such as DeepEyes~\cite{zheng2025deepeyes} and Visual Programming~\cite{gupta2023visual} delegate perception tasks to external vision models. However, such pipelines are non-differentiable and heavily reliant on the quality of off-the-shelf tools. Recent hybrid approaches attempt to achieve visually anchored reasoning by interleaving visual features with text~\cite{su2025patch}, but these typically require costly human annotation of reasoning trajectories or bounding box supervision. In contrast, our proposed CrystaL framework cultivates an intrinsic reasoning capability that is fully differentiable and self-supervised, eliminating reliance on external tools or dense annotations.

\subsection{Latent Space Reasoning}
% 超越显式的文本生成，研究重心已转向latent reasoning，即推理过程在连续的隐藏状态内部进行。在纯语言模型领域，Coconut~~\cite{meta2024coconut}和Quiet-STaR~~\cite{zelikman2024quiet}表明，模型可以通过隐式向量在输出响应前进行“思考”，而无需冗长的文字表达。这一范式如今正扩展至多模态大语言模型。Mull-Tokens~~\cite{mulltokens2025}和隐式视觉提示~~\cite{chen2025mint}引入了可学习的token，以压缩多模态上下文信息。然而，这些方法依赖于通用的语言建模目标（下一token预测），并未显式惩罚深层网络中视觉信息的丢失。因此，隐式状态往往偏向语言先验，在感知型任务中忽视了视觉证据~~\cite{ieee2025eyeswide}。除此之外，CoVT xxxxxx。与之不同，我们的工作通过引入一种visual consistency objective，显式地强制隐式推理链在输入被破坏的情况下仍保留任务相关的视觉语义，从而脱颖而出。

Moving beyond explicit text generation, research focus has shifted towards latent reasoning, wherein the inferential process unfolds within a sequence of hidden states. Within the realm of pure language models, Coconut~\cite{hao2024training} and Quiet-STaR~\cite{zelikman2024quiet} demonstrate that models can engage in `thinking' through implicit vectors prior to generating output responses, without requiring verbose textual expressions. This paradigm is now extending to multimodal large language models. For instance, Mint-CoT~\cite{chen2025mint} establishes a mapping between patch indices and visual tokens to provide direct supervision. Monet introduces a specialized dataset featuring images that describe reasoning trajectories, supervising intermediate steps by aligning token features with dataset images. Similarly, CoVT~\cite{qin2025chain} leverages external visual models to supervise the latent features of visual tokens. However, these approaches rely on generic language modeling objectives (next-token prediction) and do not explicitly penalise the utility of visual information within reasoning. In contrast, our work distinguishes itself by introducing a visual consistency objective, explicitly compelling implicit inference chains to preserve targeted visual semantics even when inputs are corrupted.

\subsection{Self-Supervised Visual Representation Learning}
% 我们的训练范式借鉴了自监督学习在计算机视觉领域的成功经验。掩码图像建模~~\cite{he2022mae, bao2021beit}和对比学习~~\cite{chen2020simclr, grill2020byol} 已证明，通过重建被破坏的输入或强调不同视角之间的一致性，可以学习到鲁棒的视觉表征。尽管这些方法用于传统编码器的预训练，但近期针对多模态大语言模型的研究已开始将这些原则应用于后训练阶段的对齐。Visual Jigsaw~~\cite{visualjigsaw2025} 利用拼图任务来增强细粒度感知能力。CapPO~~\cite{cappo2025}通过字幕一致性来减少幻觉现象。然而，这些方法通常作用于输出层面。与之不同，CrystaL将一致性正则化直接施加于隐式推理空间中。强制要求来自完整图像的隐式状态能够有效恢复模型在受损图像上的推理能力，扎根于视觉本身。

Our training paradigm draws inspiration from the success of self-supervised learning in computer vision. Masked image modeling~\cite{he2022masked} and contrastive learning~\cite{chen2020simple, grill2020bootstrap} demonstrate that robust visual representations can be learned by either reconstructing corrupted inputs. While these methods are originally designed for pretraining conventional encoders, recent studies on multimodal large language models began applying these principles during post-training alignment~\cite{li2025vitp}. For instance, visual Jigsaw~\cite{wu2025visual} enhances fine-grained perception through a jigsaw puzzle task, and CapPO~\cite{tu2025perception} reduces hallucination via caption consistency. However, these approaches typically operate at the output level. In contrast, CrystaL applies consistency regularization directly within the latent reasoning space, requiring that implicit states derived from intact images effectively restore the model’s reasoning capability on corrupted inputs—thus grounding reasoning firmly in visual content.

\begin{table}[t]
\centering
\fontsize{7.5pt}{9pt}\selectfont

% 定义更深的颜色以提高对比度
\definecolor{deepgreen}{rgb}{0.0, 0.45, 0.0}
\definecolor{deepred}{rgb}{0.7, 0.0, 0.0}

% ==== Column spacing parameter ====
\newcommand{\colspace}{4.1pt}
\setlength{\tabcolsep}{\colspace}
\renewcommand\arraystretch{1.2}

\caption{
\textbf{Key properties of visual latent reasoning methods.} In contrast to prior works like Aurora~\cite{bigverdi2025aurora}, SKILA~\cite{tong2025sketch}, LVR~\cite{li2025latent}, LIVR, and CoVT, our \textbf{CrystaL} is the only framework that simultaneously satisfies all criteria. Specifically, it eliminates the need for auxiliary modules and supplementary image labels while enabling efficient one-stage training. Symbols in \textcolor{deepgreen}{green} denote \textcolor{deepgreen}{desired} properties, while those in \textcolor{deepred}{red} denote \textcolor{deepred}{undesired} ones.
}

\begin{tabular}{l|cccccc}
\textbf{Desired Properties} 
& \textbf{Aurora} & \textbf{SKILA} & \textbf{LVR} & \textbf{LIVR} & \textbf{CoVT} & \textbf{Ours} \\
\Xhline{0.7pt}

\makecell[l]{No extra module}
& \textcolor{deepred}{\xmark} 
& \textcolor{deepred}{\xmark} 
& \textcolor{deepgreen}{\checkmark} 
& \textcolor{deepgreen}{\checkmark} 
& \textcolor{deepred}{\xmark} 
& \textcolor{deepgreen}{\checkmark} \\

\hline
\makecell[l]{No extra images \\ needed for training}
& \textcolor{deepgreen}{\checkmark} 
& \textcolor{deepred}{\xmark} 
& \textcolor{deepgreen}{\checkmark} 
& \textcolor{deepgreen}{\checkmark} 
& \textcolor{deepgreen}{\checkmark} 
& \textcolor{deepgreen}{\checkmark} \\

\hline
\makecell[l]{One-stage training}
& \textcolor{deepred}{\xmark} 
& \textcolor{deepgreen}{\checkmark}
& \textcolor{deepred}{\xmark} 
& \textcolor{deepred}{\xmark} 
& \textcolor{deepred}{\xmark} 
& \textcolor{deepgreen}{\checkmark} \\

\hline
\makecell[l]{Reason in the \\continuous space}
& \textcolor{deepred}{\xmark} 
& \textcolor{deepgreen}{\checkmark}
& \textcolor{deepgreen}{\checkmark}
& \textcolor{deepgreen}{\checkmark}
& \textcolor{deepgreen}{\checkmark}
& \textcolor{deepgreen}{\checkmark} \\
\end{tabular}
\vspace{-16pt}
\label{tab:previous-work}
\end{table}

\section{Method}
% We present CrystaL, a single-stage dual-path framework designed to instill robust latent reasoning into MLLMs. The core of our method is the Latent Crystallization process, which anchors targeted semantics into specialized latent tokens.
We propose CrystaL, a single-stage dual-path framework that instills robust latent reasoning into MLLMs via Latent Crystallization, which anchors targeted semantics into dedicated latent tokens.
We first introduce the chain of thought reasoning in Sec.~\ref{sec:3.1}. We then introduce Stochastic Image Corruption (SIC) in Sec.~\ref{sec:3.2}. Next we describe the dual-path framework used to enforce latent-level dependencies in Sec.~\ref{sec:3.3}. Finally we formulate the objective of optimization in Sec.~\ref{sec:3.4} and Sec.~\ref{sec:3.5}

\subsection{Reasoning with Chain of Thoughts}\label{sec:3.1}

\begin{figure*}[htbp]
    \centering
    \includegraphics[width=1\linewidth]{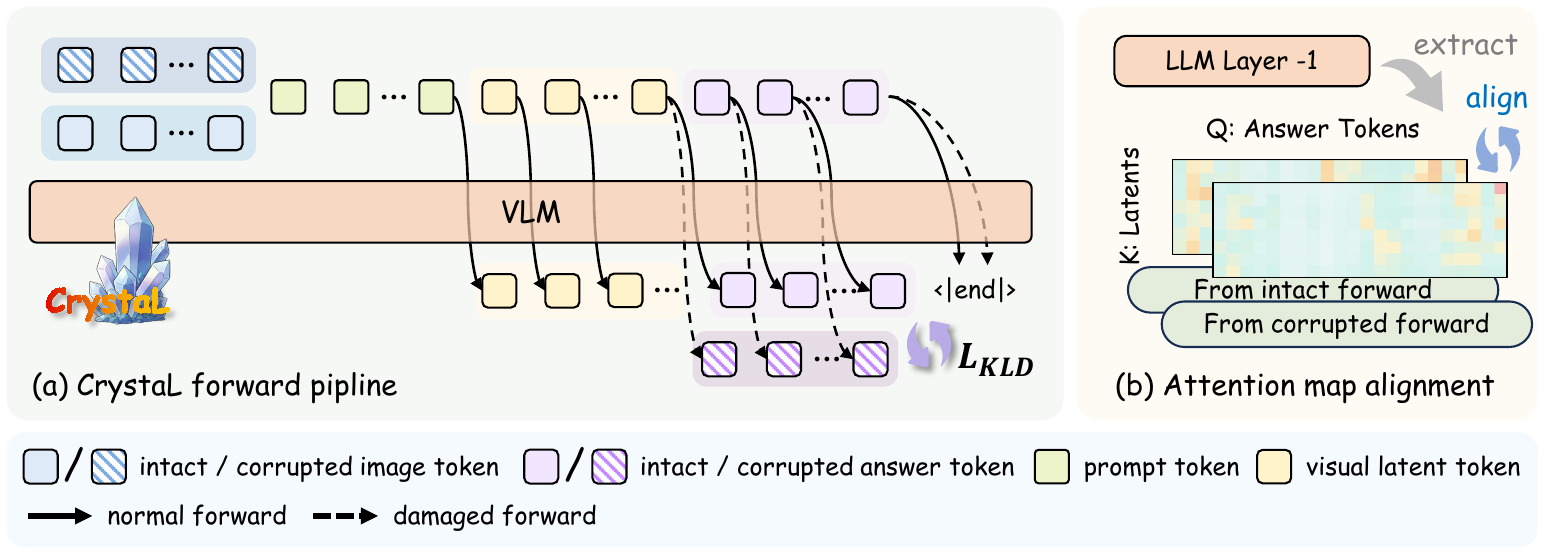}
    \caption{\textbf{An overview of CrystaL.} (a) Starting by corrupting the raw image, we construct two types of visual tokens from vision encoder. Then both the $S_{int}$ and $S_{cor}$ in Sec.~\ref{sec:3.3} are fed into the model to compute the $\mathbf{P}_{int}$ and $\mathbf{P}_{cor}$. But the hidden states of corrput forward (dashed arrow) indeed come from the the intact forward (solid arrow). For the objective function, we adopt a combination of cross entropy loss and alignment loss, the attention map alignment is illustrated in (b). }
    \label{fig:main}
\end{figure*}

In the context of Multimodal Large Language Models (MLLMs), let $\mathcal{M}$ denote the model. Given a visual input $I$ and a textual query $Q$, the model generates an output sequence $Y = \{Y_0, Y_1, \dots, Y_n\}$ token by token. The generation process is formulated as:
\begin{equation}
    P_\mathcal{M}(Y | I, Q) = \prod_{i} P_\mathcal{M}(Y_i | Y_{<i}, I, Q),
\end{equation}
where $Y_{<i}$ represents the sequence of tokens predicted prior to step $i$. In standard Fine-tuning for Visual Question Answering (VQA), the objective is to maximize the likelihood of the ground-truth answer $P_\mathcal{M}(Ans | I, Q)$.

To enhance the model's complex reasoning capabilities, we incorporate the Chain-of-Thought (CoT) paradigm. Under this setting, the objective function is based on a joint probability distribution:
\begin{equation}
    P_\mathcal{M}(Ans, CoT | I, Q),
\end{equation}
where $CoT$ represents a sequence of intermediate reasoning steps. The introduction of $CoT$ effectively decomposes the complex mapping from visual-textual inputs to the final answer, allowing the model to allocate more computational steps to internal logic before arriving at a conclusion.

While conventional CoT sequences consist of discrete textual tokens, the concept of \textbf{Latent CoT} has been proposed. In this formulation, reasoning steps are represented as continuous latent variables within the model's hidden space. Our framework primarily supervises the representations in the latent space during training to preserve high-dimensional semantic nuances that discrete tokens might fail to capture.
%Although these latent states can be mapped back to discrete token IDs,

\subsection{Visual Corruption as Information Bottleneck}\label{sec:3.2}
A fundamental challenge in training latent reasoning models is to ensure the visual latent tokens used, since raw visual tokens often contain sufficient cues for the model to answer the question, it tends to skip these latent reasoning paths. This implicitly hinders the emergence of genuine visual latent reasoning, as the model arrives at answers without effectively engaging the continuous hidden states. To address this, we introduce the \textbf{Stochastic Image Corruption (SIC)} module. SIC acts as a generalized operator $\mathcal{C}$ that maps the high-fidelity image $I$ into a degraded observation $I_{cor}$:
\begin{equation}
    I_{cor} = \mathcal{C}(I; \eta, \mathcal{S}), \quad \eta \sim p(H),
\end{equation}
where $\mathcal{S}$ denotes the corruption mode and $\eta$ is the stochastic intensity sampled from a prior distribution $p(H)$. To explore the resilience of latent reasoning, we consider a diverse library of corruption primitives $\mathcal{S} \in \{\mathcal{G}, \mathcal{K}, \mathcal{V}, \mathcal{N}\}$:

\begin{itemize}
    \item \textbf{Spectral Decay ($\mathcal{G}$):} Gaussian-based low-pass filtering that suppresses high-frequency textures while preserving global topology.
    \item \textbf{Spatial Discontinuity ($\mathcal{K}$):} Random cropping, masking, or Jigsaw shuffling that disrupts the geometric coherence of visual evidence.
    \item \textbf{Chromatic Distortion ($\mathcal{V}$):} Jittering color distributions to isolate semantic shape from pixel-level intensity by decoupling invariant structural geometry from transient global photometric fluctuations.
    \item \textbf{Stochastic Noise ($\mathcal{N}$):} Additive Gaussian or impulse noise to the input tensors.
\end{itemize}

\subsection{Dual-Path Latent Reasoning Framework}\label{sec:3.3}

The model processes the multimodal query through two parallel streams illustrated in Fig.~\ref{fig:main}. During training, the \textbf{causal attention mechanism} allows the computation of output logits for the entire sequence in a single forward pass. Specifically, for an input containing an image $I$ and text sequence $text = \{t_1,t_2,\dots,t_L\}$ of length $L$, the model $\mathcal{M}$ can generate a sequence of hidden states $\mathbf{H} = \{h_1, \dots, h_L\}$, which are subsequently mapped to the predicted probability distributions (logits) over the vocabulary $\mathcal{V}$ at each position. So in the training stage the labels directly come from the shifted sequence. As we add several pad tokens to shape the $CoT = \{intermediate~steps,<visual~tokens>\}$, the sequence is like $\{I, Q, CoT, Ans\}$, where $Ans$ is the ground-truth answer. Then we process the sequence in the dual-path framework.

\paragraph{The Intact Path.} 
This path receives the high-fidelity image $I_{int} = I$. The input sequence is formulated as $S_{int} = \{I_{int}, Q, CoT, Ans\}$. As the ``teacher'' stream, it leverages complete visual evidence to compute the latent states value of visual tokens $ \mathbf{T}_{lat} = \{T_{1},T_{2},\dots,T_{l} \}$ ($l$ represents the number of layers of the language model) to provide the raw information of an intact sample such that the corrupted path will borrow the latents to conduct the ``crytalization" operation. And the intact path naturally generates the logits $\mathbf{P}_{int}$ for the computation of cross entropy loss and work as a reference of the corrupted path.

\begin{figure}[t!]
    \centering
    \includegraphics[width=1\linewidth]{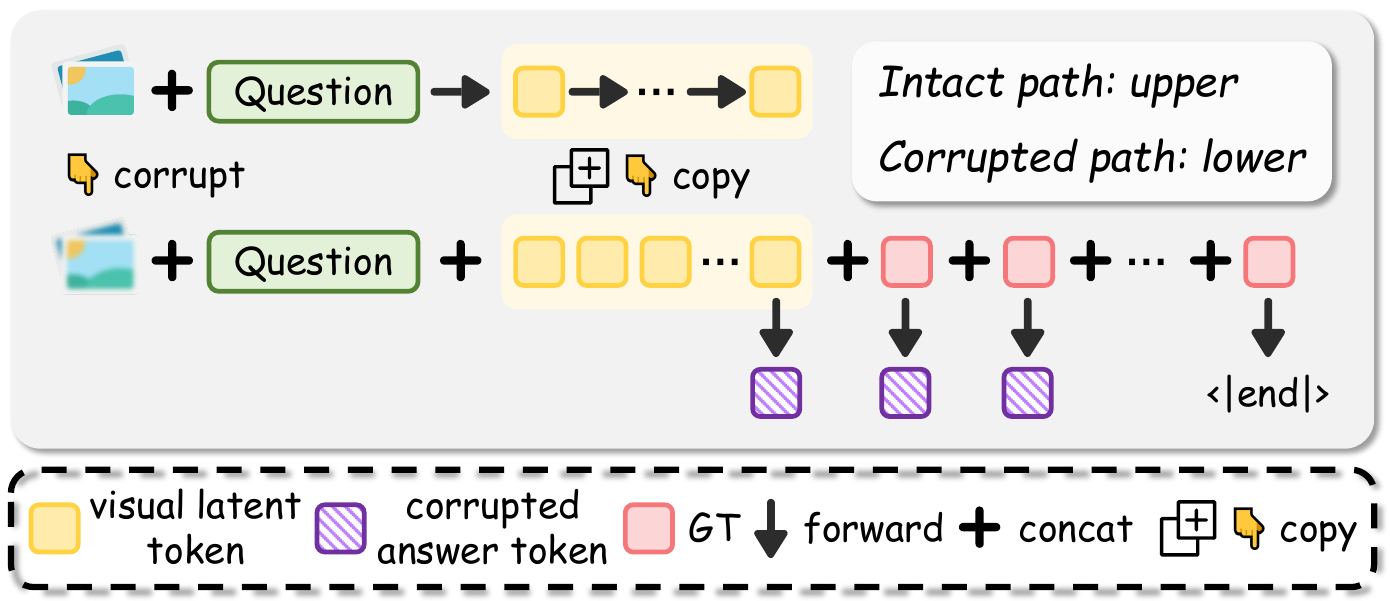}
    \caption{\textbf{Detailed illustration of visual latent token copying.} For the forward process, the autoregressive hidden states of visual latent tokens in the corrupted path is copied from the intact path.}
    \label{fig:replicate}
\end{figure}

\paragraph{The Corrupted Path.}
This path operates on the perturbed image $I_{cor} = \mathcal{C}(I)$. The corresponding input sequence $S_{cor} = \{I_{cor}, Q, CoT, Ans\}$. We enforce a \textbf{latent-level dependency} by replacing the latent representations at each decoder layers in the corrupted path with corresponding latents in $\mathbf{T}_{lat}$ derived from the intact path. The model then generates the logits $\mathbf{P}_{cor}$ for the perturbed sequence.

By constructing the dual-path framework, we bypass the difficulty of labeling data by hand to supervise the visual tokens. Because the $\mathbf{T}_{lat}$ with complementary information from intact path will facilitate the corrupted path, which enforce the visual tokens to work as indispensible ingredient of reasoning. The supervision from the next token loss will naturally work on the visual tokens.

\subsection{Alignment Objectives}\label{sec:3.4}
As the above dual-path framework constructs the base of ``crystalization", the hidden states $\mathbf{T}_{lat}$ has been copied from the intact path to the corrupted path. Then we need to supervise the model to motivate the $\mathbf{T}_{lat}$ to absorb the robust feature of the image.

\begin{table*}[t]
\setlength{\tabcolsep}{10pt}
\centering \footnotesize
\renewcommand\arraystretch{1.05}
\caption{\textbf{Comparison of model performance across diverse visual benchmarks.} The results highlight CrystaL's superior capabilities comparing with baselines including Qwen2.5-VL-7B~\cite{bai2025qwen2}.}
\begin{tabular}{@{}lccccccccc@{}}
\toprule
\multicolumn{1}{c}{} & \multicolumn{2}{c}{\textbf{CVBench}} & \multicolumn{2}{c}{\textbf{HRBench}} &  &  &  &  &  \\ \cmidrule(lr){2-5}
\multicolumn{1}{c}{\multirow{-2}{*}{\vspace{4pt}\textbf{Method}}} & 2D & 3D & 4K & 8K & \multirow{-2}{*}{\vspace{4pt}\textbf{BLINK}} & \multirow{-2}{*}{\vspace{4pt}\textbf{RWQA}} & \multirow{-2}{*}{\vspace{4pt}\textbf{V*}} & \multirow{-2}{*}{\vspace{4pt}\textbf{POPE}} & \multirow{-2}{*}{\vspace{4pt}\textbf{Average}} \\ \midrule
\multicolumn{10}{c}{\cellcolor[HTML]{F2F2F2}\textbf{Closed-source Methods}} \\
Claude-4-Sonnet & - & - & 32.3 & 22.7 & 39.6 & 63.7 & 15.2 & - & - \\
GPT-4o & - & - & 50.6 & 46.7 & 63.0 & 69.7 & 42.9 & - & - \\ \midrule
\multicolumn{10}{c}{\cellcolor[HTML]{F5F9FD}\textbf{Open-source Methods}} \\
Qwen2.5-VL-7B & 75.0 & 73.3 & 68.6 & 64.9 & 55.7 & 68.6 & 76.4 & 86.5 & 71.1 \\
CoVT~\cite{qin2025chain} & {\underline{75.1}} & \textbf{84.7} & 71.0 & {\underline{69.4}} & {\underline{56.0}} & \textbf{71.6} & 78.0 & 84.6 & 73.8 \\
LVR~\cite{li2025latent} & - & - & 69.6 & 66.1 & 52.5 & 67.7 & 81.7 & - & - \\
Vision-R1~\cite{huang2025vision} & ~~8.0 & 37.6 & 64.8 & 57.0 & 51.0 & ~~9.4 & 80.1 & 77.8 & 48.2 \\
LIVR~\cite{li2025latent_livr} & 73.9 & 82.7 & {\underline{72.4}} & 69.0 & 56.3 & 69.2 & 79.1 & \textbf{89.5} & \underline{74.0} \\
SKILA~\cite{tong2025sketch} & - & - & 72.0 & 66.5 & \textbf{56.7} & 70.5 & \textbf{84.3} & 87.3 & - \\ \midrule
\multicolumn{10}{c}{\cellcolor[HTML]{F5FFFA}\textbf{Our Method}} \\
\textbf{CrystaL} & \textbf{76.6} & {\underline{84.4}} & \textbf{73.4} & \textbf{71.1} & 55.9 & {\underline{70.6}} & {\underline{82.7}} & {\underline{88.7}} & \textbf{75.4} \\
$\Delta$(vs Qwen2.5-VL-7B) & {\color[HTML]{32CB00} \textbf{+1.6}} & {\color[HTML]{32CB00} \textbf{+11.1}} & {\color[HTML]{32CB00} \textbf{+4.8}} & {\color[HTML]{32CB00} \textbf{+6.2}} & {\color[HTML]{32CB00} \textbf{+0.2}} & {\color[HTML]{32CB00} \textbf{+2.0}} & {\color[HTML]{32CB00} \textbf{+6.3}} & {\color[HTML]{32CB00} \textbf{+2.2}} & {\color[HTML]{32CB00} \textbf{+4.3}} \\ \bottomrule
\end{tabular}
\label{tab:main}
\end{table*}

\paragraph{Predictive Distribution Consistency ($\mathcal{L}_{kl}$).} 
To effectively anchor the reasoning process within the latent space, we supervise the corrupted path using the high-fidelity outputs of the intact path as a reference. This is achieved through a consistency-driven alignment objective applied to the predictive distributions.

Specifically, the objective is to conduct a logit-level loss between $\mathbf{P}_{int}$ and $\mathbf{P}_{cor}$, targeting at the tokens corresponding to the ground-truth $Ans$. This targeted alignment compels the latent reasoning states to ``crystallize'' into targeted visual representations that are functionally indispensable for the final generative process.

To ensure that the model maintains robust decision-making despite visual degradation, we minimize the Kullback-Leibler (KL) divergence between the output probability distributions of the two paths. Specifically, this loss is calculated over the tokens corresponding to the ground-truth answer $A = \{a_1, \dots, a_M\}$, forcing the corrupted path to emulate the predictive confidence of the intact path:
\begin{equation}
    \mathcal{L}_{kl} = \frac{1}{M} \sum_{i \in \Omega_A} D_{KL} \left( \mathcal{P}(\hat{y}_i | S_{int}) \parallel \mathcal{P}(\hat{y}_i | S_{cor}) \right),
\end{equation}
where $\Omega_A$ denotes the set of indices for the answer tokens, and $\mathcal{P}(\hat{y}_i | \cdot)$ represents the predicted probability distribution over the vocabulary $\mathcal{V}$ at the $i$-th position. By optimizing this objective, we compel the model to recover the targeted semantics from the transferred latent tokens $\mathbf{T}_{lat}$, bridging the gap caused by the input corruption.

\paragraph{Mechanistic Consistency ($\mathcal{L}_{attn}$).} 
While predictive consistency aligns the final outputs, it does not inherently guarantee that both path follow the same inferential pattern. We then enforce consistency within the internal attention mechanism. Specifically, we constrain the attention weights assigned by the answer tokens to the visual latent tokens. Let $\mathbf{A} \in \mathbb{R}^{H \times M \times K}$ denote the aggregated attention maps across $H$ heads, representing the attention scores from $M$ answer tokens to $K$ latent tokens. We minimize the squared Frobenius norm of the discrepancy between the two paths:
\begin{equation}
    \mathcal{L}_{attn} = \mathbb{E}_{l \in \mathcal{L}_{sub}} \left\| \mathbf{A}_{int}^{(l)} - \mathbf{A}_{cor}^{(l)} \right\|_F^2,
\end{equation}
where $\mathcal{L}_{sub}$ represents a selected subset of decoder layers. By optimizing $\mathcal{L}_{attn}$, we ensure that even when fine-grained visual details are neutralized, the model’s reasoning behavior—reflected in its localized attention over the latent space—remains stable and firmly grounded in the semantic representations encoded by the visual latent tokens distilled from the original image.

\subsection{Overall Training Objective}\label{sec:3.5}

To facilitate latent crystallization while preserving the fundamental instruction-following and generative capabilities of the MLLM, we integrate the proposed alignment objectives with the cross entropy loss into a unified \textit{single-stage} optimization framework.

\paragraph{Cross Entropy Loss $\mathcal{L}_{ce}^{int}$ and $\mathcal{L}_{ce}^{cor}$.}
A cross-entropy loss on the corrupted path, denoted as $\mathcal{L}_{ce}^{cor}$, encourages the model to seek information from the visual latent tokens when the visual input is corrupted and thus unavailable, thereby enabling the latent representations to encode meaningful visual information. In order to keep the model output the sequence in the format of standard chain of thought, a cross entropy loss in the intact path $\mathcal{L}_{ce}^{int}$ is also introduced. $\mathcal{L}_{ce}^{int}$ will then enforce the model make use of the visual latent tokens to keep reasoning ability in the vanilla mode where the image is intact.

% The total training loss is formulated as a weighted combination of the generative and consistency-driven terms:
% \begin{equation}
%     \label{eq:total_loss}
%     \mathcal{L}_{\text{total}} = \mathcal{L}_{\text{ce}}^{\text{int}} + \mathcal{L}_{\text{ce}}^{\text{cor}} + \lambda_{1} \mathcal{L}_{\text{kl}} + \lambda_{2} \mathcal{L}_{\text{attn}},
% \end{equation}
% where $\mathcal{L}_{\text{ce}}^{\text{int}}$ and $\mathcal{L}_{\text{ce}}^{\text{cor}}$ denote the standard next-token prediction losses (cross-entropy) computed on the intact and corrupted paths, respectively. The coefficients $\lambda_{1}$ and $\lambda_{2}$ serve as hyperparameters to balance the consistency constraints with the generative objectives.

The total training loss is formulated as a combination of the generative and consistency-driven terms:
\begin{equation}
    \label{eq:total_loss}
    \mathcal{L}_{\text{total}} = \mathcal{L}_{\text{ce}}^{\text{int}} + \mathcal{L}_{\text{ce}}^{\text{cor}} +  \mathcal{L}_{\text{kl}} + \mathcal{L}_{\text{attn}},
\end{equation}
where $\mathcal{L}_{\text{ce}}^{\text{int}}$ and $\mathcal{L}_{\text{ce}}^{\text{cor}}$ denote the standard next-token prediction losses (cross-entropy) computed on the intact and corrupted paths, respectively.

Under this unified formulation, the model is incentivized not only to generate coherent textual responses but also to \textit{actively} encode task-essential visual semantics into the specialized latent tokens $\mathbf{T}_{lat}$. As a result, $\mathbf{T}_{lat}$ functions as a repository of \textbf{semantic invariants} that capture high-level visual evidence capable of sustaining complex reasoning even under severely degraded perceptual conditions. This objective ensures the emergence of a latent reasoning path that is both \textbf{consistent across diverse visual views} and \textbf{deeply grounded} in the underlying visual logic, providing a scalable solution to the challenge of misalignment between supervision signals and visual latent CoT.

\section{Experiment}

\subsection{Experimental Setup}

\paragraph{Training Configurations}
We fine-tune our model using LoRA~\cite{hu2022lora} with a rank $r=16$ and an alpha $\alpha=32$. The learning rates are set to $2 \times 10^{-4}$ for LoRA adapters. All experiments are conducted on a workstation equipped with $4 \times \text{NVIDIA A40}$ GPUs.

\paragraph{Benchmarks.}
We evaluate all models using VLMEvalKit~\cite{duan2024vlmevalkit}, a standardized evaluation framework for multimodal large language models. 
To comprehensively assess the visual perception and reasoning capabilities of CrystaL, we conduct experiments across multiple representative benchmarks covering complementary evaluation aspects. We adopt RealWorldQA (RWQA)~\cite{zhang2024realworld}, BLINK~\cite{fu2024blink}, V*Bench (V*)~\cite{wu2024v}, and HRBench (HR4K and HR8K)~\cite{wang2025divide}. These benchmarks focus on fine-grained visual understanding, compositional reasoning, and high-resolution image comprehension, encompassing challenging real-world scenarios such as complex visual layouts and detailed spatial relations. To further examine the robustness of CrystaL against multimodal hallucination, we evaluate our model on POPE~\cite{li2023pope}, which is specifically designed to diagnose hallucinated visual claims in MLLMs.

% \paragraph{Baseline}
% To comprehensively evaluate our approach, we benchmark CrystaL against a carefully selected suite of baselines. First, to isolate the impact of our proposed reasoning framework from the training data itself, we establish a Direct SFT baseline, which is a text-only SFT variant trained on the exact same dataset and with identical parameters as CrystaL. We then introduce several state-of-the-art models. For comparison against alternative visual supervision techniques, we include CoVT, which highlights their perception performance improvement. We also compare LVR\cite{li2025latent}, a two-stage latent reasoning method combining SFT and GRPO; Vision-R1\cite{huang2025vision}, a text reasoning model got through supervised finetuning and reinforcement learning; SKILA, a visual latent paradigm for unified multimodal reasoning; LIVR ,a two-stage method using non-causal mask to construct visual bottleneck.

\paragraph{Baselines.}
We compare CrystaL with several representative methods that explore latent or explicit reasoning in multimodal models. 
\textbf{LVR} is a two-stage framework that performs supervised fine-tuning followed by reinforcement learning to induce latent visual reasoning.
\textbf{SKILA} introduces a sketch-in-latent paradigm, aiming to elicit unified multimodal reasoning by encoding structured visual abstractions into latent representations.
\textbf{LIVR} constructs an implicit visual bottleneck via non-causal attention masking, encouraging the model to rely on latent visual cues rather than direct pixel-level features.
\textbf{Vision-R1}~\cite{huang2025vision} enhances multimodal reasoning ability through a combination of supervised fine-tuning and reinforcement learning, focusing on reasoning behavior rather than directly supervising by visual grounding.

\subsection{Quantitative analysis}

Comprehensive results are summarized in Table~\ref{tab:main}, where the data reveals that CrystaL sets a new state-of-the-art across most benchmarks, achieving a leading average score of 75.4\%. Specifically, CrystaL demonstrates a significant advantage in high-resolution perception, outperforming the baseline Qwen2.5vl by 4.8\% and 6.2\% on the challenging HRBench 4K and 8K tracks.

Furthermore, while models like LIVR and SKILA show competitive results in specific hallucination or video benchmarks (e.g., POPE and VStarBench), CrystaL maintains the most balanced performance profile. The comparative degradation of Vision-R1 in CVBench-2D and RWQA highlights the robustness of our architectural design in maintaining multimodal reasoning consistency.

\begin{figure}[t!]
    \centering
    \includegraphics[width=0.8\linewidth]{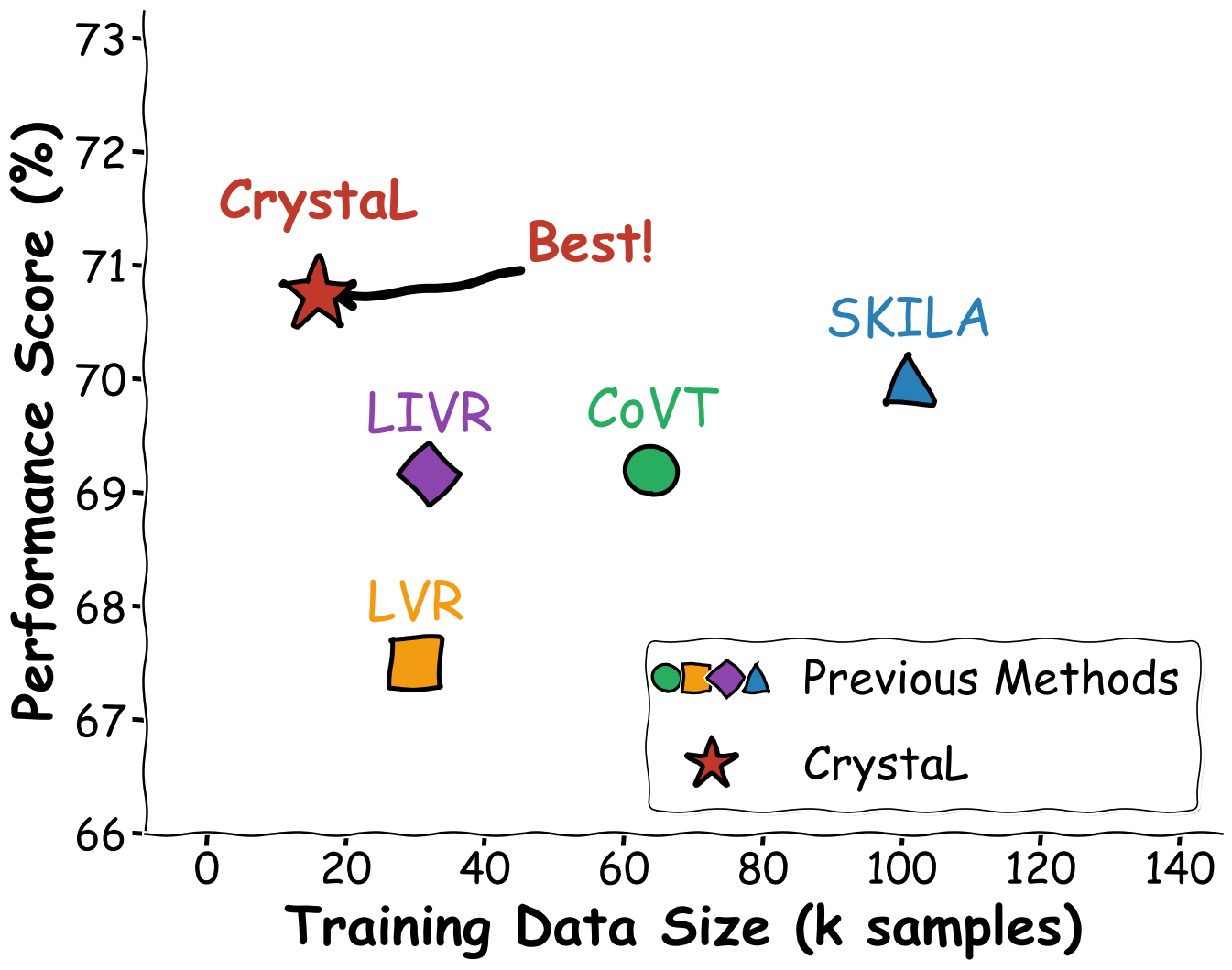}
    \caption{\textbf{Comparison of performance and training data size.} Our method achieves superior performance while utilizing significantly fewer samples than baselines, demonstrating exceptional data efficiency.}
    \vspace{-6pt}
    \label{fig:bubble_plot}
\end{figure}
\textbf{Data Efficiency.} We evaluate the training efficiency of our framework by comparing the performance against various baselines on the intersection of benchmarks, including BLINK, HRBench, RealWorld-QA, and VStarBench. As illustrated in Fig.~\ref{fig:bubble_plot}, our method achieves a state-of-the-art accuracy using only 16k training samples. Notably, this outperforms SKILA, which requires 100k samples. This comparison highlights that our latent reasoning mechanism enables the model to achieve superior multimodal understanding with over $6\times$ fewer data, demonstrating a significant advantage in data-scarce scenarios.

\subsection{Ablation Study}

\begin{table}[t!]
\setlength{\tabcolsep}{2pt}
\centering
\fontsize{8.5pt}{9.5pt}\selectfont
\caption{\textbf{ Ablation on the quantity and content of latent reasoning tokens.} The upper section denotes configurations using same tokens, while the lower section utilizes diverse tokens.}
\begin{tabular}{@{}lcccccc|c@{}}
\toprule
\multicolumn{1}{c}{Quantity} & $\text{CV}_{2D}$ & $\text{CV}_{3D}$ & BLINK & $\text{HR}_{4K}$ & $\text{HR}_{8K}$ & RWQA & Avg \\ \midrule
\multicolumn{8}{c}{\cellcolor[HTML]{F5F9FD}Baseline} \\
Qwen2.5-VL & 75.0 & 73.3 & 55.7 & 68.6 & 64.9 & 68.6 & 67.7 \\ \midrule
\multicolumn{8}{c}{\cellcolor[HTML]{FFFBF0}Identical latent token} \\
4 tokens & 74.7 & 74.7 & 52.9 & 73.8 & 71.0 & 68.0 & 69.2 \\
8 tokens & 75.7 & 83.7 & 54.4 & \textbf{74.4} & \textbf{72.0} & 69.3 & 71.6 \\
16 tokens & 72.3 & 80.6 & 52.3 & 71.5 & 68.1 & 69.7 & 69.1 \\ \midrule
\multicolumn{8}{c}{\cellcolor[HTML]{FDF2EA}Diverse latent token} \\
4 tokens & 75.1 & 82.9 & 55.3 & 73.8 & 70.8 & 69.8 & 71.3 \\
8 tokens & \textbf{76.6} & \textbf{84.4} & \textbf{55.9} & 73.4 & 71.1 & 70.6 & \textbf{72.0} \\
16 tokens & 74.0 & 72.6 & 53.9 & 74.3 & 71.4 & \textbf{71.6} & 69.6 \\ \bottomrule
\end{tabular}
\label{tab:ablation-num}
\end{table}

% \textbf{Ablation on token numbers and type.} We investigate the impact of latent token configurations on the reasoning capacity of CrystaL. As the Table~\ref{tab:ablation-num} shows, our results reveal two key insights: first, employing diverse tokens consistently outperforms the use of identical tokens, suggesting that semantic variety in the latent space is crucial for capturing complex multimodal dependencies. Second, we observe a non-monotonic relationship between performance and token count, where a length of 8 tokens achieves the optimal balance, surpassing both shorter (4) and longer (16) sequences.
\textbf{Ablation on token numbers and type.} This section provides a detailed analysis that elucidates the impact of latent token configurations on the overall capacity of CrystaL. As Table~\ref{tab:ablation-num} shows, our results reveal two key insights: first, employing diverse tokens consistently outperforms the use of identical tokens, a finding which underscores that semantic variety in the latent space is crucial for capturing complex multimodal dependencies. Second, we observe a non-monotonic relationship between performance and token count, where a length of 8 tokens achieves the optimal balance, surpassing both shorter (4) and longer (16) sequences in both settings.

\begin{table}[t!]
\setlength{\tabcolsep}{1.1pt}
\centering
\fontsize{7.5pt}{9pt}\selectfont
\caption{\textbf{ Ablation on alignment strategy.} Both the $\mathcal{L}_{kl}$ and $\mathcal{L}_{attn}$ play a critical role in our method. And the type of $\mathcal{L}_{attn}$ is also important. Here, ``Cor." means ``corrupted".}
\begin{tabular}{@{}lccccccc@{}}
\toprule
\multicolumn{1}{c}{Strategy} & $\text{CV}_{2D}$ & $\text{CV}_{3D}$ & BLINK & $\text{HR}_{4K}$ & $\text{HR}_{8K}$ & \multicolumn{1}{c|}{RWQA} & Avg \\ \midrule
\multicolumn{8}{c}{\cellcolor[HTML]{FFFBF0}How does the intact path assist corrupted path?} \\
Qwen2.5-VL & 75.0 & 73.3 & 55.7 & 68.6 & 64.9 & \multicolumn{1}{c|}{68.6} & 67.7 \\
+ Cor. CE & 74.7 & 83.3 & 54.6 & 73.0 & 70.3 & \multicolumn{1}{c|}{69.9} & 71.0 \\
+ Cor. CE + KL & 75.2 & 83.6 & 54.1 & \textbf{75.0} & 71.1 & \multicolumn{1}{c|}{\textbf{71.1}} & 71.7 \\
+ Cor. CE + Att. Align & 76.2 & 84.0 & \textbf{57.1} & 72.6 & 70.4 & \multicolumn{1}{c|}{70.3} & 71.8 \\
+ All & \textbf{76.6} & \textbf{84.4} & 55.9 & 73.4 & \textbf{71.1} & \multicolumn{1}{c|}{70.6} & \textbf{72.0} \\ \midrule
\multicolumn{8}{c}{\cellcolor[HTML]{FDF2EA}Attention map alignment strategy} \\
Answer to All (KL) & 73.6 & 79.5 & 53.9 & 74.8 & 69.4 & \multicolumn{1}{c|}{69.5} & 70.1 \\
Answer to Latents (KL) & \textbf{76.6} & \textbf{84.4} & \textbf{55.9} & 73.4 & 71.1 & \multicolumn{1}{c|}{\textbf{70.6}} & \textbf{72.0} \\
Answer to Latents (MSE) & 76.3 & 83.1 & 54.0 & \textbf{73.8} & \textbf{71.4} & \multicolumn{1}{c|}{70.2} & 71.4 \\ \bottomrule
\end{tabular}
\label{tab:ablation-loss}
\end{table}

\textbf{Abalation on alignment strategy.} To demonstrate the effectiveness of our alignment strategy, we conduct an ablation study on the loss functions. As shown in Table~\ref{tab:ablation-loss}, the model yields the lowest performance when optimized solely with cross-entropy loss. Incorporating KL-loss or Alignment-loss into the objective consistently improves results, with the performance reaching its peak when all components are combined in the Full configuration.

Furthermore, we investigate how the specific formulation of alignment affects model performance. While adopting a standard MSE-based alignment provides marginal gains, it lacks the flexibility to capture complex cross-modal dependencies. Notably, we found that applying KL-type alignment to the entire attention map (from answer to both image and latent tokens) actually degrades performance. The model achieves its best performance only when the alignment is restricted to the attention from answer tokens (as queries) to latent tokens (as keys).

\begin{table}[t!]
\centering
\fontsize{8.3pt}{10pt}\selectfont
\setlength{\tabcolsep}{1.6pt}
\renewcommand\arraystretch{0.95}
\newcommand{\pos}[1]{\textcolor{green!60!black}{#1}}
\newcommand{\negval}[1]{\textcolor{red!70!black}{#1}}
\caption{\textbf{ Ablation on corruption strategy.} The Gaussian blur behaves best in all the corruption strategies, while random mask is also useful for learning visual latents.}
\begin{tabular}{lcccccc|c}
\toprule
Strategy & $\text{CV}_{2D}$ & $\text{CV}_{3D}$ & BLINK & $\text{HR}_{4K}$ & $\text{HR}_{8K}$ & RWQA & Avg\\
\midrule
Gaussian Blur    &76.6 & 84.4 & 55.9 & 73.4 & 71.1 & 70.6 & 72.0\\
Random Mask &76.3 & 84.7 & 55.2 & 71.9 & 69.4 & 68.4 & 71.0 \\
Jigsaw Shuffling & 73.6 & 82.5 & 54.0 & 71.4 & 69.0 & 65.2 & 69.3\\
Gaussian Noise   &74.9 & 81.8 & 54.1 & 72.0 & 69.8 & 69.9& 70.4\\
Colour Distortion    &74.8 & 80.3 & 54.6 & 72.3 & 70.4 & 70.6 &70.5\\
\bottomrule
\end{tabular}
\label{tab:corruption}
\end{table}

\textbf{Abalation on corruption strategy.} We evaluate different corruption strategies $\mathcal{C}$ in the proposed SIC module, including Gaussian blur, random masking, colour distortion, jigsaw shuffling, additive Gaussian noise. As shown in Table~\ref{tab:corruption}, Gaussian blur consistently achieves the best performance across all benchmarks, while other corruption types lead to inferior results.

We hypothesize that spatial perturbations (e.g., Jigsaw or random masking) often introduce artificial edge artifacts that lead to a distribution shift in the visual encoder's feature map, potentially hindering the model's basic recognition. But Gaussian blur strikes the best balance between information suppression and semantic preservation, making it the most suitable corruption strategy in our framework. Consequently, we adopt Gaussian-based corruption as the default setting in other experiments.

\begin{figure}[t!]
    \centering
    \includegraphics[width=1\linewidth]{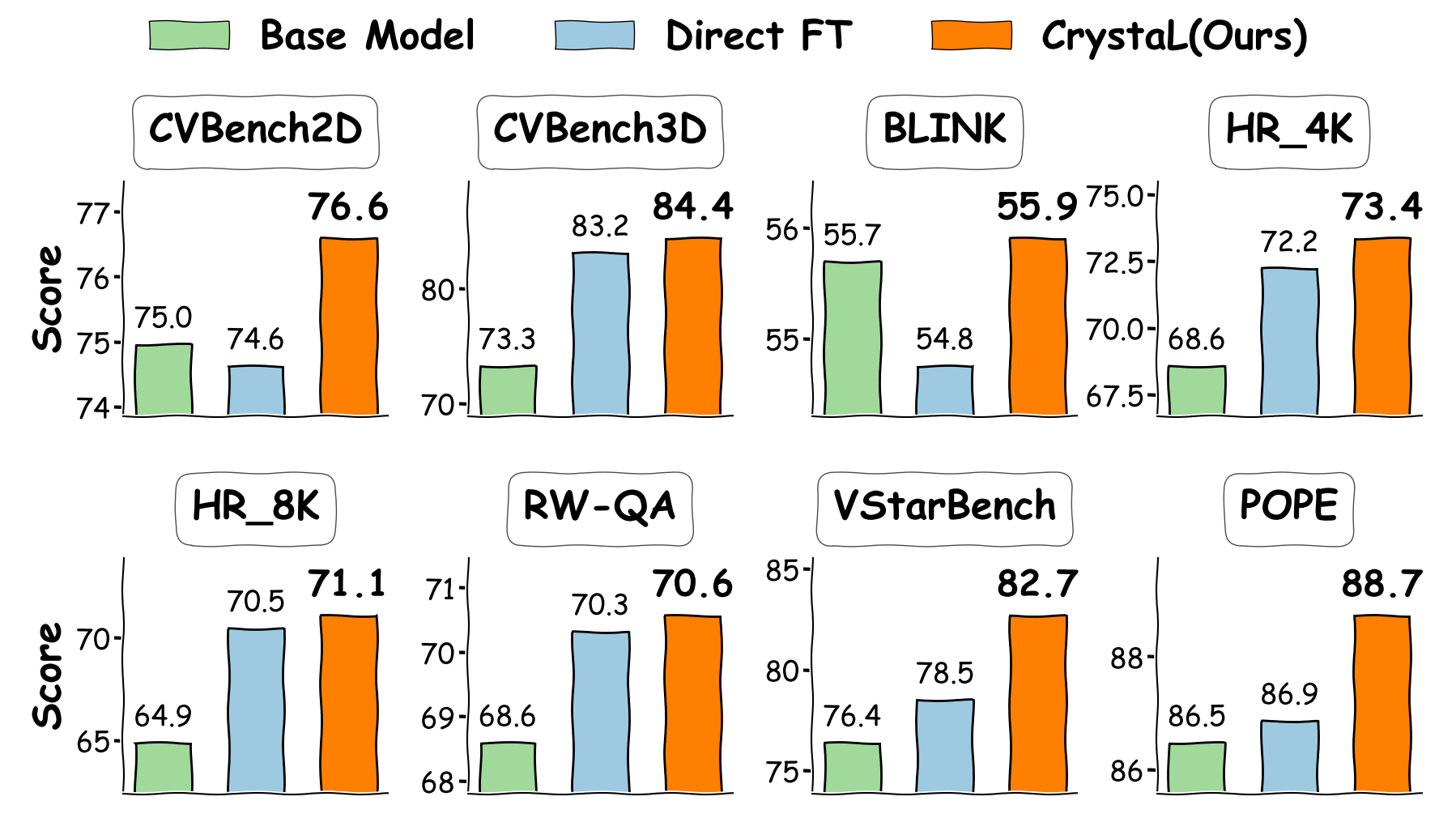}
    \caption{\textbf{Direct Finetune vs CrystaL.} CrystaL outperforms direct finetuning’capabilities across all the tasks.}
    \vspace{-6pt}
    \label{fig:direct_hist}
\end{figure}

\textbf{Ablation on direct finetune}
To isolate the contribution of our latent reasoning mechanism, we conduct an ablation study comparing CrystaL with a direct fine-tuning (Direct FT) baseline. As shown in Fig.~\ref{fig:direct_hist}, although Direct FT improves upon the base model, it consistently underperforms compared to CrystaL across all eight benchmarks. Specifically, in perception-intensive tasks such as CVBench2D and VStarBench, CrystaL achieves absolute gains of 2.7\% and 4.2\% over Direct FT, respectively. This performance gap demonstrates that simply updating model weights via standard supervised fine-tuning is insufficient for complex multimodal tasks. Instead, our latent thinking process provides a more effective pathway for the model to internalize and utilize dense visual evidence for inference.

\subsection{Qualitative Analysis}

\begin{figure}[t!] 
    \centering
    \includegraphics[width=0.95\linewidth]{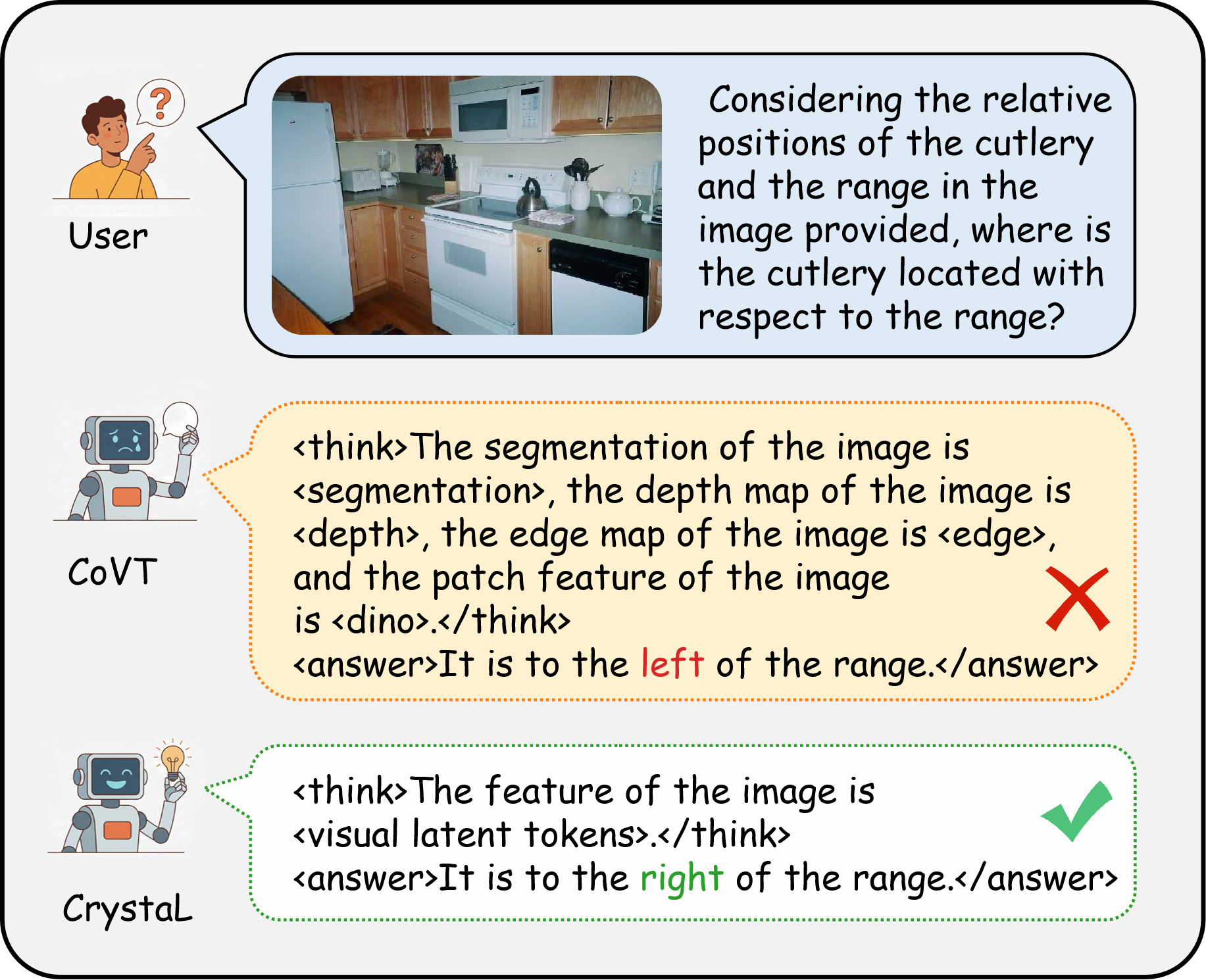}
    % \vspace{-8pt}
    \caption{\textbf{Qualitative comparison on CVBench-2D.} We illustrate the responses from CoVT and CrystaL to a question regarding spatial-relational reasoning between objects.}
    \vspace{-6pt}
    \label{fig:vqa_example}
\end{figure}
To demonstrate the enhanced perceptual capabilities afforded by our CrystaL framework, we present a case study from the CVBench-2D dataset where CrystaL outperforms CoVT as Fig.~\ref{fig:vqa_example} shows. Notably, even with the integration of SAM, DINO,  Depth Anything~\cite{yang2024depth} for structural alignment, CoVT struggles in relations between objects while CrystaL yields the correct answer.

\section{Conclusion}
In this paper, we introduced CrystaL, a novel single-stage training framework for visual latent reasoning. By employing a dual-path architecture, CrystaL enables visual latent tokens to be naturally supervised without relying on external modules or auxiliary images. Our experimental results demonstrate that CrystaL not only achieves superior performance across various benchmarks but also establishes a more efficient paradigm for visual reasoning. This work provides a scalable and streamlined approach for the development of multimodal large language models.

\newpage
\section*{Impact Statement}
This paper presents work whose goal is to advance the field of Machine
Learning. There are many potential societal consequences of our work, none
which we feel must be specifically highlighted here.

% \newpage

\bibliography{example_paper}
\bibliographystyle{icml2026}

%%%%%%%%%%%%%%%%%%%%%%%%%%%%%%%%%%%%%%%%%%%%%%%%%%%%%%%%%%%%%%%%%%%%%%%%%%%%%%%
%%%%%%%%%%%%%%%%%%%%%%%%%%%%%%%%%%%%%%%%%%%%%%%%%%%%%%%%%%%%%%%%%%%%%%%%%%%%%%%
% APPENDIX
%%%%%%%%%%%%%%%%%%%%%%%%%%%%%%%%%%%%%%%%%%%%%%%%%%%%%%%%%%%%%%%%%%%%%%%%%%%%%%%
%%%%%%%%%%%%%%%%%%%%%%%%%%%%%%%%%%%%%%%%%%%%%%%%%%%%%%%%%%%%%%%%%%%%%%%%%%%%%%%
\newpage
\appendix
\onecolumn
\section{Appendix}

\subsection{Experiment Details}

\paragraph{Training Datasets.} 
To ensure a fair comparison and maintain consistency with prior work, we \textbf{directly adopt} the training suite established by~\cite{qin2025chain} without any additional modification. The dataset comprises vision-centric subsets from LLaVA-OneVision~\cite{li2024llava-onevision}, TallyQA~\cite{acharya2019tallyqa}, and ADE20K-Depth~\cite{zhou2017scene}, following the exact configuration described in the original study.

\paragraph{Training Configurations.} 
We fine-tune our model using LoRA~\cite{hu2022lora} with a rank $r=16$ and an alpha $\alpha=32$. The learning rates are set to $2 \times 10^{-4}$ for LoRA adapters. All experiments are conducted on a workstation equipped with $4 \times \text{NVIDIA A40}$ GPUs.

\paragraph{Settings of baselines}

We directly obtained the reported results of SKILA and LVR from their respective papers. For \textbf{SKILA}, we adopt the Unified Reasoning Model (SKILA) rather than SKILA-V. The results for \textbf{LVR} are taken from the experimental reports in the SKILA paper.For \textbf{CoVT}, we downloaded the CoVT-seg-dino-depth model and conducted local evaluation following the official setup.For \textbf{LIVR}, we follow the training protocol described in the original paper, performing a two-stage training process with 4K steps for each stage.
\subsection{Examples of conversation on different tasks}
\begin{figure}[htbp] % 学术论文通常首选 [t] (top)
    \centering
    \includegraphics[width=\linewidth]{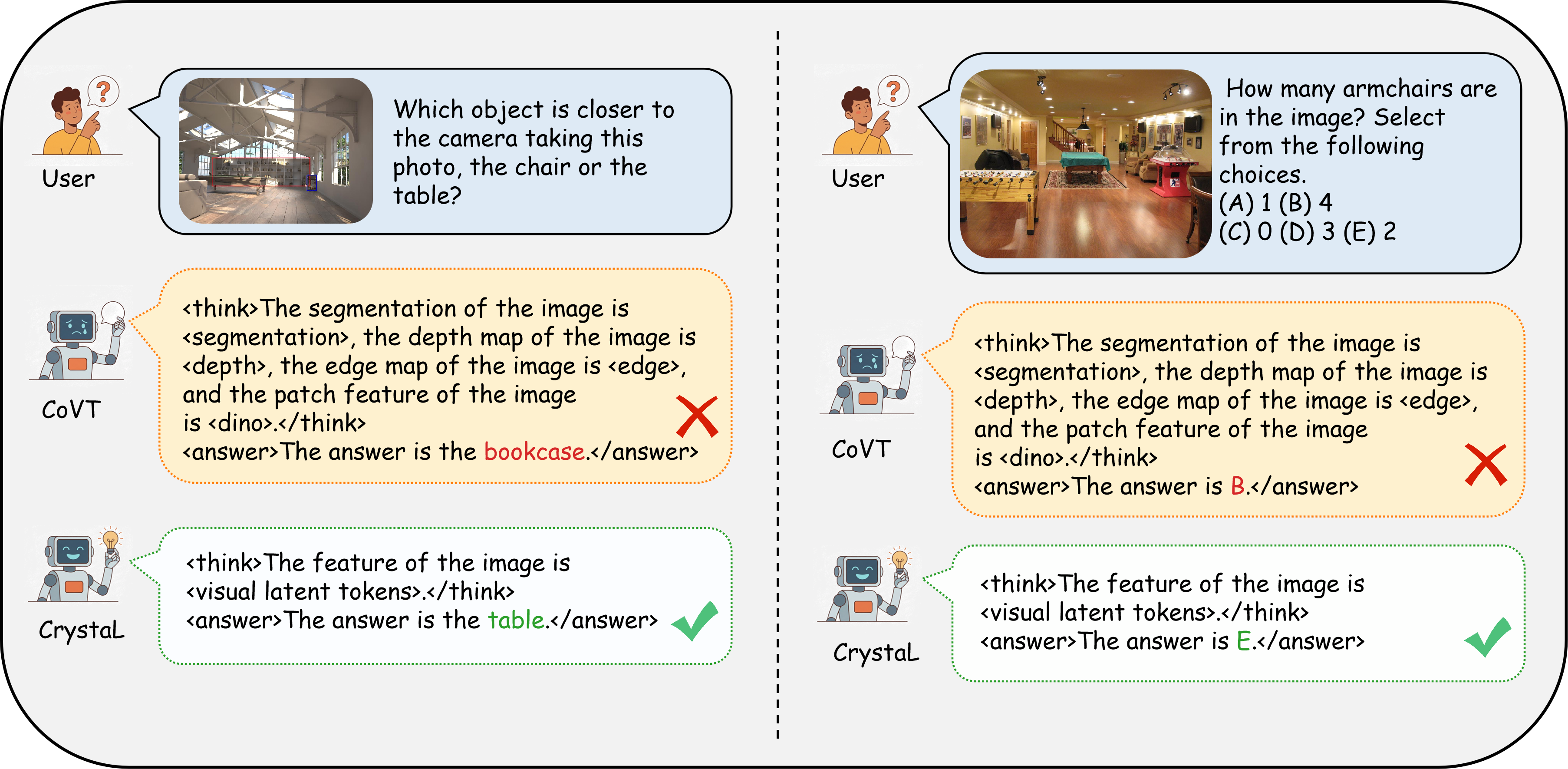}
    \caption{\textbf{Results on counting and depth}}
    \label{fig:vqa_append1}
\end{figure}

\begin{figure}[htbp] % 学术论文通常首选 [t] (top)
    \centering
    \includegraphics[width=\linewidth]{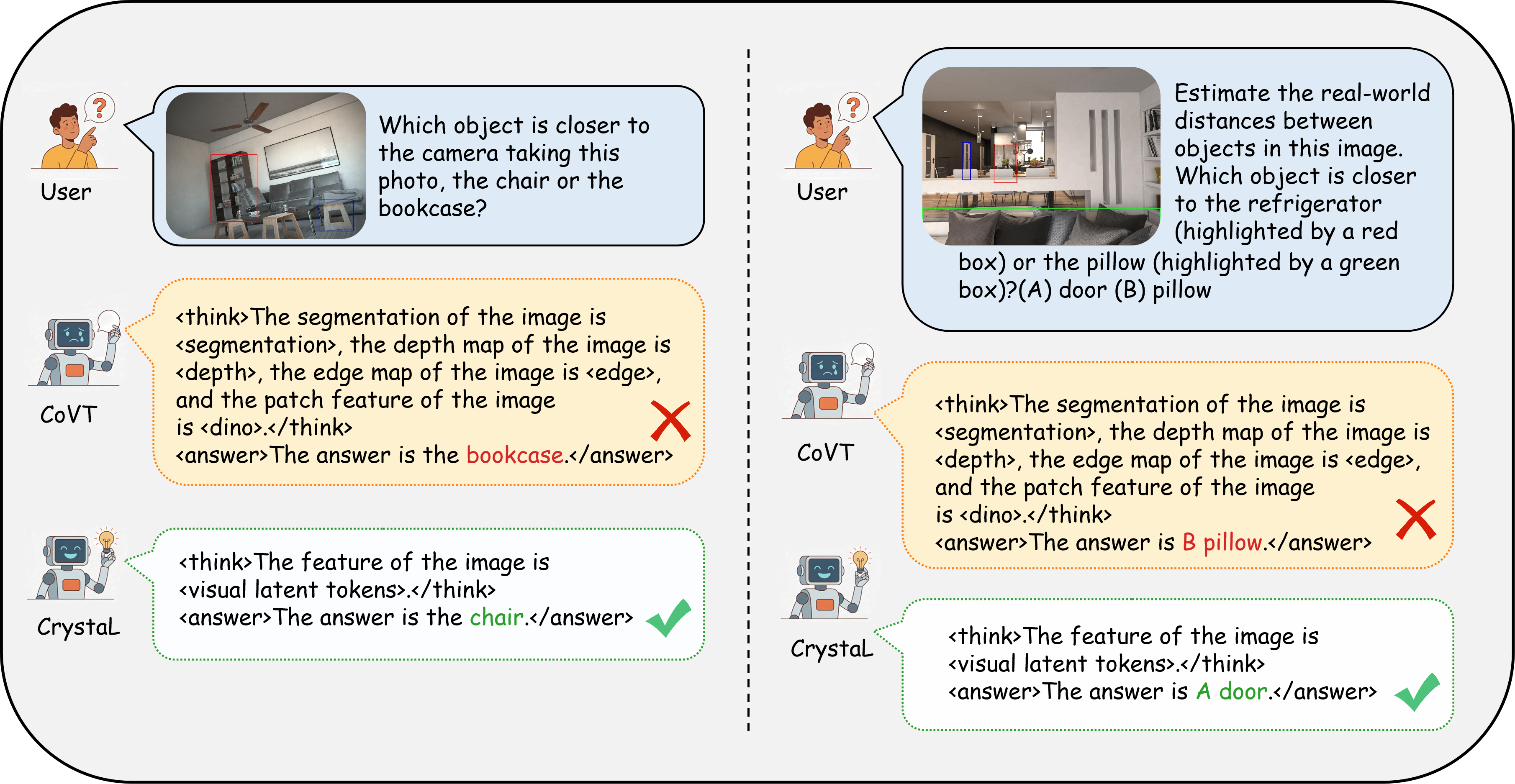}
    \caption{\textbf{Results on relation and distance.}}
    \label{fig:vqa_append2}
\end{figure}

We provide more question and answer examples in Fig.~\ref{fig:vqa_append1} and Fig.~\ref{fig:vqa_append2}.

\subsection{Demonstration on corrupted examples}
\begin{figure}[htbp] % 学术论文通常首选 [t] (top)
    \centering
    \includegraphics[width=0.7\linewidth]{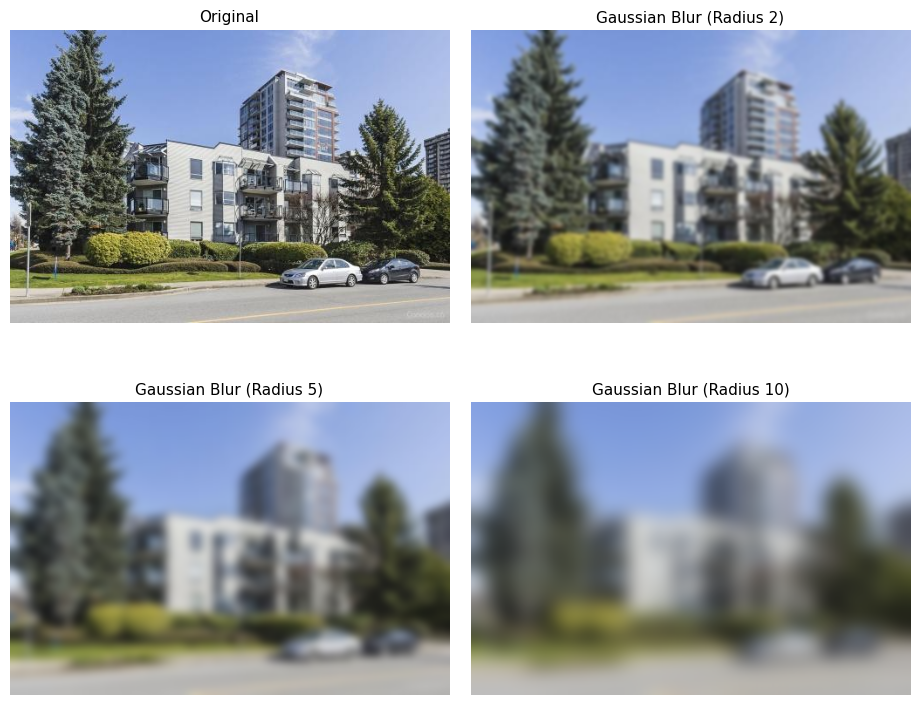}
    \caption{\textbf{Gaussian blur variations.} We illustrate the multi-scale blurring responses applied to the input samples, showing the original image and three levels of Gaussian smoothing ($r=2, 5, 10$).}
    \label{fig:blur_example}
\end{figure}

\begin{figure}[htbp] % 学术论文通常首选 [t] (top)
    \centering
    \includegraphics[width=0.7\linewidth]{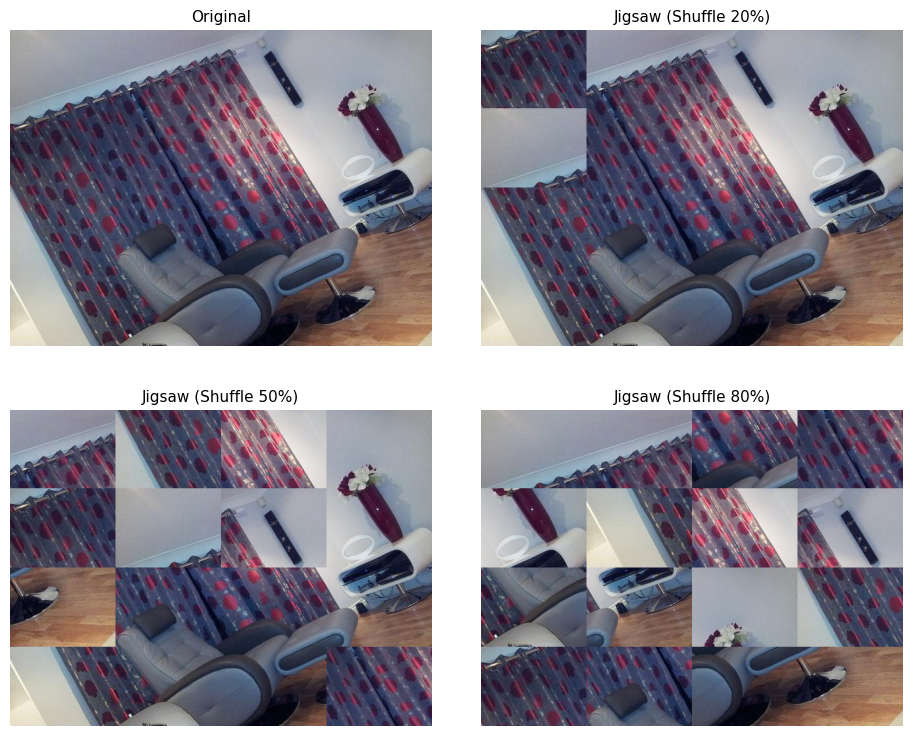}
    \caption{\textbf{Jigsaw.} We apply Jigsaw augmentation to the input samples to challenge the model's ability to capture global structural consistency. Three levels of shuffling ratios ($\rho \in \{20\%, 50\%, 80\%\}$) are illustrated to demonstrate increasing spatial complexity.}
    \label{fig:jigsaw_example}
\end{figure}

\begin{figure}[htbp] % 学术论文通常首选 [t] (top)
    \centering
    \includegraphics[width=0.7\linewidth]{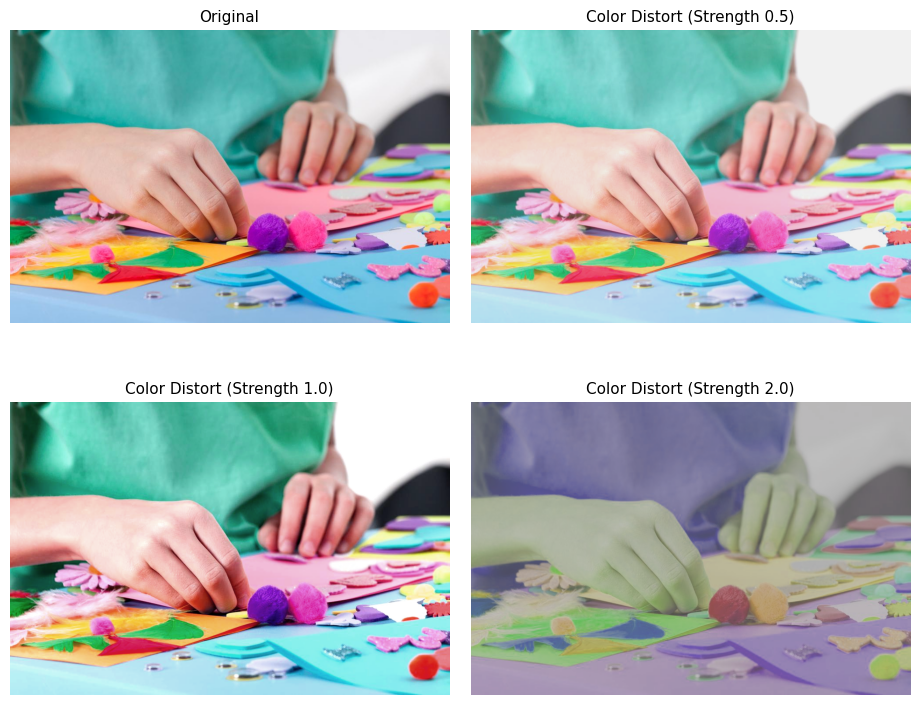}
    \caption{\textbf{Colour Distort.} We illustrate the visual variations generated by the color jittering module by modulating brightness, contrast, saturation, and hue with increasing intensities ($s \in \{0.5, 1.0, 2.0\}$).}
    \label{fig:color_example}
\end{figure}

\begin{figure}[htbp] % 学术论文通常首选 [t] (top)
    \centering
    \includegraphics[width=0.7\linewidth]{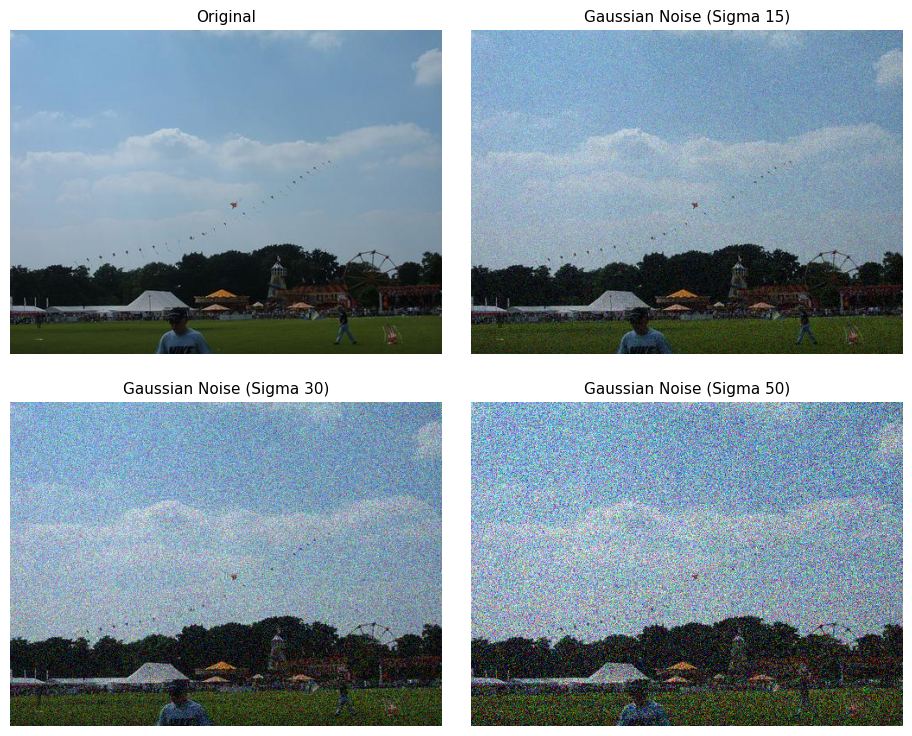}
    \caption{\textbf{Gaussian Noise.} We illustrate the original image and its corrupted versions subjected to additive Gaussian noise with varying standard deviations ($\sigma \in \{15, 30, 50\}$). }
    \label{fig:noise_example}
\end{figure}

\begin{figure}[htbp] % 学术论文通常首选 [t] (top)
    \centering
    \includegraphics[width=0.7\linewidth]{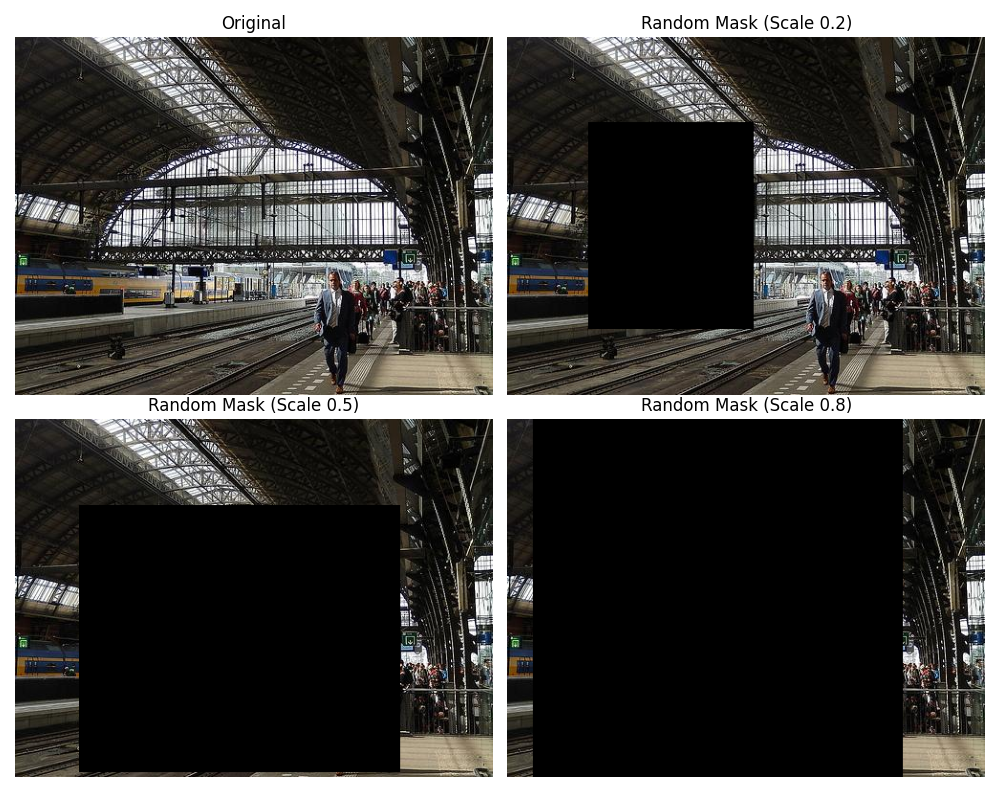}
    \caption{\textbf{Random mask.} We demonstrate the original image alongside three variants of Random Erasing at different scales ($s \in \{0.2, 0.5, 0.8\}$).}
    \label{fig:cut_example}
\end{figure}

The illustration of five kinds of corruption $\mathcal{C}$ is in ~Fig.~\ref{fig:blur_example},~\ref{fig:jigsaw_example},~\ref{fig:noise_example},~\ref{fig:color_example},~\ref{fig:cut_example}

\subsection{Attention map during training}
\begin{figure}[htbp] % 学术论文通常首选 [t] (top)
    \centering
    \includegraphics[width=\linewidth]{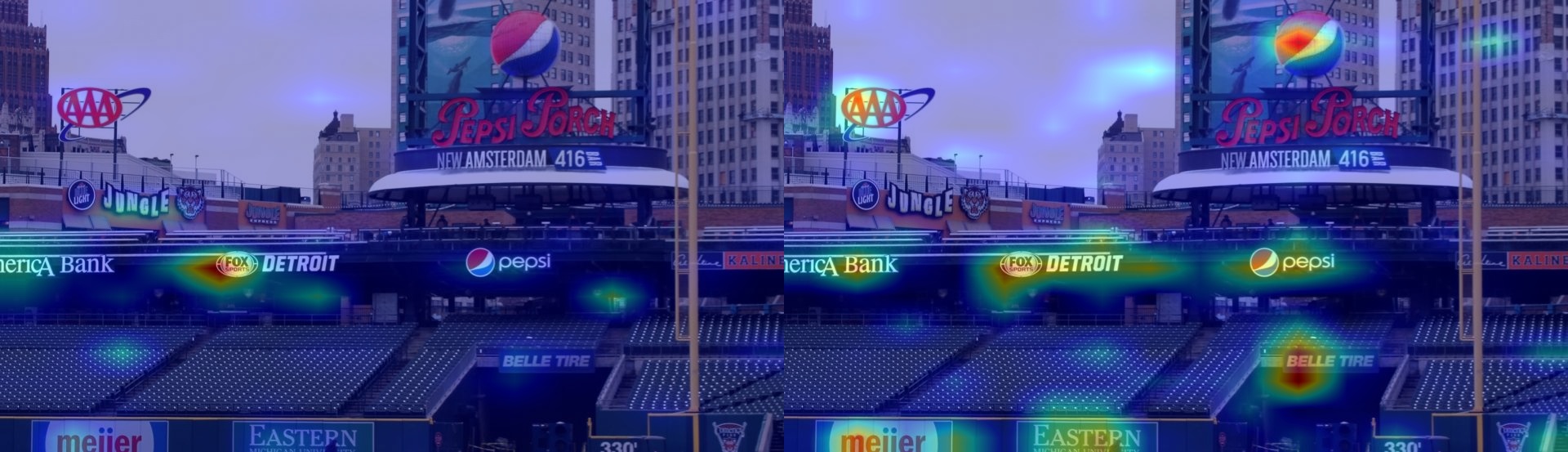}
    \caption{\textbf{Attention differences.} The intact path attends to detroit while the corrupted path attends to all the labels.}
    \label{fig:12_attn}
\end{figure}

\begin{figure}[htbp] % 学术论文通常首选 [t] (top)
    \centering
    \includegraphics[width=\linewidth]{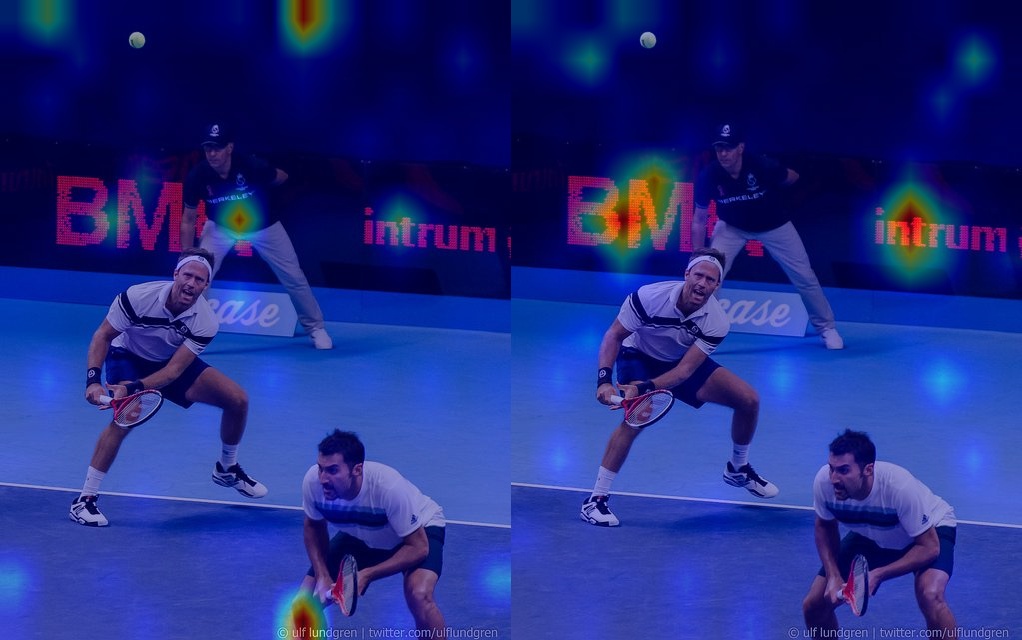}
    \caption{\textbf{Attention differences.} The intact path attends to random positions while the corrupted path attends to the text behind the man.}
    \label{fig:13_attn}
\end{figure}

\begin{figure}[htbp] % 学术论文通常首选 [t] (top)
    \centering
    \includegraphics[width=\linewidth]{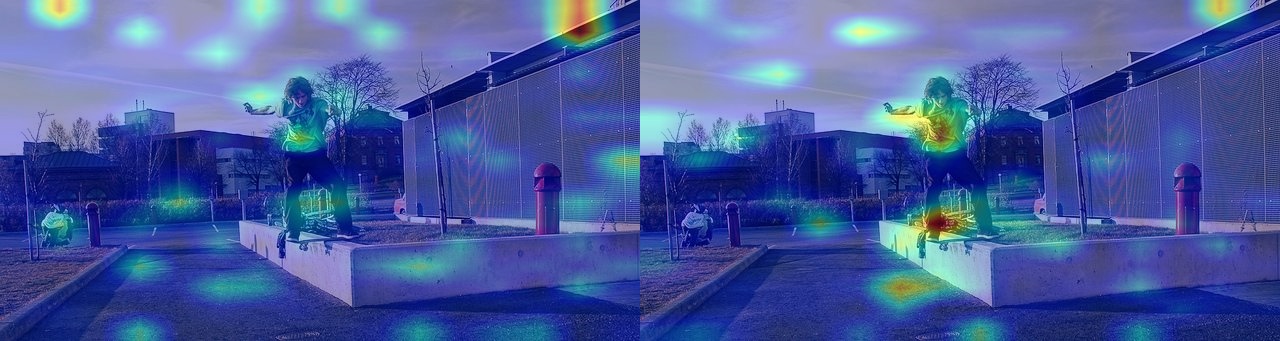}
    \caption{\textbf{Attention differences.} The intact path attends to random positions while the corrupted path attends to the man.}
    \label{fig:14_attn}
\end{figure}

To analyze how the two paths work, we visualize the attention maps during training in Fig.~\ref{fig:12_attn},~\ref{fig:13_attn},~\ref{fig:14_attn}. The left part corresponds to the intact path, while the right one corresponds to the corrupted path.

\end{document}